\documentclass[letterpaper]{article} 
\usepackage{aaai23}  
\usepackage{times}  
\usepackage{helvet}  
\usepackage{courier}  
\usepackage[hyphens]{url}  
\usepackage{graphicx} 
\usepackage{natbib}  
\usepackage{caption} 
\usepackage{multirow}
\usepackage{comment}
\usepackage{amsmath}
\usepackage{amsfonts}
\usepackage{amssymb}
\usepackage{amsbsy}
\usepackage{xcolor}
\usepackage{mathtools}
\usepackage{algorithm}
\usepackage{algorithmic}
\usepackage{pdflscape}
\usepackage{newfloat}
\usepackage{listings}
\DeclareCaptionStyle{ruled}{labelfont=normalfont,labelsep=colon,strut=off} 
\floatstyle{ruled}
\newfloat{listing}{tb}{lst}{}
\floatname{listing}{Listing}
\title{A Bayesian approach to quantifying uncertainties and improving generalizability in traffic prediction models}

\author{
    Agnimitra Sengupta\textsuperscript{\rm 1}, 
    Sudeepta Mondal\textsuperscript{\rm 2}\equalcontrib,
    Adway Das\textsuperscript{\rm 3}\equalcontrib, S. Ilgin Guler\textsuperscript{\rm 4}\equalcontrib\\
}
\affiliations{
    \textsuperscript{\rm 1,3,4}Department of Civil \& Environmental Engineering, The Pennsylvania State University,\\
    University Park, PA 16802, USA\\
    \textsuperscript{\rm 2}Department of Mechanical Engineering, The Pennsylvania State University,\\
    University Park, PA 16802, USA\\
    \textsuperscript{\rm 1}avs6897@psu.edu, \textsuperscript{\rm 2}sudeepta.mondal2@rtx.com,
    \textsuperscript{\rm 3}amd7293@psu.edu, \textsuperscript{\rm 4}sig123@psu.edu
}

\usepackage{bibentry}

\begin{document}
\maketitle
\begin{abstract}
Deep-learning models for traffic data prediction can have superior performance in modeling complex functions using a multi-layer architecture. However, a major drawback of these approaches is that most of these approaches do not offer forecasts with uncertainty estimates, which are essential for traffic operations and control. Without uncertainty estimates, it is difficult to place any level of trust to the model predictions, and operational strategies relying on overconfident predictions can lead to worsening traffic conditions. In this study, we propose a Bayesian recurrent neural network framework for uncertainty quantification in traffic prediction with higher generalizability by introducing spectral normalization to its hidden layers. In our paper, we have shown that normalization alters the training process of deep neural networks by controlling the model's complexity and reducing the risk of overfitting to the training data. This, in turn, helps improve the generalization performance of the model on out-of-distribution datasets. Results demonstrate that spectral normalization improves uncertainty estimates and significantly outperforms both the layer normalization and model without normalization in single-step prediction horizons. This improved performance can be attributed to the ability of spectral normalization to better localize the feature space of the data under perturbations. Our findings are especially relevant to traffic management applications, where predicting traffic conditions across multiple locations is the goal, but the availability of training data from multiple locations is limited. Spectral normalization, therefore, provides a more generalizable approach that can effectively capture the underlying patterns in traffic data without requiring location-specific models.
\end{abstract}

\section{Introduction}
Efficient traffic management including traffic control and congestion mitigation rely on accurate short-term traffic predictions into the future (i.e., ranging from 5 min to 1 hr). Different traffic control strategies, such as ramp metering or detour suggestions, are reliant on precise traffic forecasting in the short and near-short future. Precise forecasting across free-flow and congested traffic states is often challenging due to the uncertain and chaotic nature of transportation systems. To address the challenges of short-term traffic prediction, various modeling approaches have been proposed in the literature, including classical statistical models and machine learning (ML)-based methods. These models aim to capture the complex and dynamic nature of traffic flow, which arises due to the interaction of multiple factors such as traveler behavior, road networks, traffic incidents, and weather conditions \cite{zhu2017prediction,zhu2017road}. Several statistical parametric techniques including historical average algorithms, smoothing techniques, autoregressive integrated moving average (ARIMA) \cite{ahmed1979,levin1980} were proposed to model the temporal fluctuations in traffic.
However, parametric methods are often limited in their performances due the specific assumptions on the functional relationship. As a result, non-parametric approaches that do not specify a specific functional form are a promising alternative to model traffic patterns with greater transferability and robustness across datasets \cite{smith1997, clark2003}. 
For example, methods like nearest neighbors \cite{smith2002}, support vector machine \cite{mingheng2013}, and Bayesian network \cite{sun2006} have been used in short-term traffic forecasting.

Deep learning (DL) approaches have demonstrated superior performance in predicting future traffic conditions compared to parametric and other non-parametric approaches. Neural networks (NN) are capable of approximating a complex relationship using a series of non-linear transformations of the input data. Therefore, researchers have considered using NN to model temporal patterns in traffic data \cite{chang1995,innamaa2000,dia2001}. 
In particular, recurrent neural network (RNN) \cite{rumelhart1986, hochreiter1997} and its variants like long short-term memory (LSTM) are exclusively designed to handle temporal data by accounting for the temporal correlation in data. For example, \cite{van2002} used RNN architecture for travel time prediction, whereas, LSTM based architectures have been used for short-term travel speed prediction \cite{ma2015, cui2020} and predicting traffic under extreme conditions \cite{yu2017}.

Despite the success of DL in traffic forecasting, most models suffer from a significant drawback -- the models output deterministic predictions of the short-term traffic variable (e.g., flow, density or speed) and their performance are assessed based on the prediction error \cite{karlaftis2011}. These models do not include uncertainty estimates for the prediction, which are crucial for comprehending the model's limitations and for further application in an active traffic management system. Uncertainty estimation is a crucial aspect of traffic management, as it enables decision-makers to make informed predictions by providing a measure of confidence in model outputs. Traffic data is inherently noisy and complex, making it challenging to make accurate predictions with point estimates alone. Failure to account for uncertainty can lead to sub-optimal traffic management decisions that do not consider the range of possible outcomes, potentially leading to increased congestion, delays, or safety risks. 

Prediction uncertainty can be either \textit{aleatoric}, which arises from the intrinsic randomness in the data generation process, or \textit{epistemic}, which is uncertainty in the model specification \cite{kiureghian2009}. Aleatoric uncertainty may arise due to sensor malfunction and general randomness in traffic demand, and hence is irreducible with higher data volumes. To the contrary, epistemic uncertainty can be reduced by collecting more data and developing more complex models allowing the exploration of under-represented regions in the training data space.

Limited research on uncertainty quantification have been performed for traffic prediction. For example, stochastic model based on a seasonal autoregressive integrated moving average (SARIMA) model combined with a generalized autoregressive conditional heteroscedasticity (GARCH) model have been utilized to generate traffic flow forecasts and prediction intervals \cite{guo2007data}. Further, for real-time ITS applications, adaptive Kalman filters with time-varying process variances were used to handle the stochastic traffic time series models \cite{guo2010,guo2014}. 
Prediction intervals for bus travel time or freeway travel time have been developed using a simple 2-layer NN with Bayesian and delta methods to address the complexity associated with the underlying traffic processes and the uncertainty associated with the data used to infer travel time \cite{Khosravi2011}. 

DL models with deterministic outputs that are trained using stochastic gradient descent can efficiently converge to a local minima with a high probability depending on the chosen initial points, where all local minima attain similar loss values \cite{lee2016gradient}. However, to account for model variance, uncertainty quantification in DL models using variational inference explores model specifications in the neighborhood of one such local minimum. This is achieved by fitting probability distributions on model parameters, rather than using deterministic weights. 
For instance, \cite{zhu2017deep} used an encoder-decoder architecture for feature learning, followed by a LSTM prediction module with Bayesian inference to estimate model uncertainty, arising due to model specifications in the neighborhood of one local minimum. 
To the contrary, deep-ensembling techniques that combine inferences from multiple local minima (instead of one in variational inference) have also been used for uncertainty quantification in traffic studies, highlighting some advantages over variational inference. For example, \cite{mallick2022} used a Bayesian hyperparameter optimization to select a set of high-performing configurations and fit a generative model to capture the joint distributions of the hyperparameters. Then, using samples of hyperparameters generated from that distribution, an ensemble of models were trained to estimate model uncertainty. Conversely, \cite{qian2022uncertainty} combined variational inference and deep ensembling techniques by using Monte Carlo dropout and adaptive weight averaging to find multiple local minima (and, model parameters), which are suitably combined for estimating uncertainty.
Although these studies presented different Bayesian approaches to modeling uncertainty, models were primarily trained to predict on data with same or similar distributions, and hence lack generalizability. This is significant since, the model's performance on dataset with different distributions, referred to as \textit{out-of-distribution} dataset, can vary significantly among these local minima. 

This study focuses on examining the generalizability of models in terms of performance under data perturbation. Ideally, for a model that converges to a \textit{flat} local minima, the loss function remains relatively unchanged even when subjected to data perturbations. This characteristic suggests a desirable level of generalizability \cite{hochreiter1997flat}. 
Past research have used adversarial training \cite{goodfellow2014explaining} to achieve insensitivity to training data perturbation, which is not always sufficient for achieving insensitivity to test data perturbation. The goal of this study is to develop an uncertainty quantification model for traffic flow prediction that can generalize well on datasets with different distributions. 
Specifically, we use dropout and a normalization technique, namely spectral normalization to establish a Bayesian recurrent neural network that has better generalizability.

The novel contributions of this paper are:
\begin{enumerate}
    \item introducing spectral normalization to the dropout-based uncertainty quantification framework to enhance the generalizability of uncertainty quantification models
    \item conducting a systematic analysis to examine the impact of normalization on model training and understanding how it improves the model's generalizability.
\end{enumerate}

The structure of this paper is organized as follows. First, we provide a background on the DL model and Bayesian approaches used for uncertainty estimation, as well as the normalization techniques employed in this study. We then describe the data and present a comparison of the performance of our proposed models with a baseline model across various prediction tasks. Finally, based on the results, we provide concluding remarks.

\section{Background}
In this section, we provide an overview of the DL forecasting model i.e., long short-term memory (LSTM) and the uncertainty quantification method used in this paper. 

\subsection{Long short-term memory}
Feed-forward neural network architectures are not explicitly designed to handle sequential data. A class of DL approaches, recurrent neural network (RNN), uses a feedback mechanism where the output from a previous time step is fed as an input to the current time step such that information from the past can propagate into future states. This feedback mechanism preserves the temporal correlation and makes it suitable to capture the temporal evolution of traffic parameters. However, RNNs are incapable of handling the long-term dependencies in temporal data due to the vanishing gradient problem \cite{hochreiter1998}. Long short-term memory (LSTM) \cite{hochreiter1997}, a type of RNN, consists of memory cells in its hidden layers and several gating mechanisms, which control information flow within a cell state (or, memory) to selectively preserve long-term information. 

\begin{figure}[!htb]
\centering
\includegraphics[width=0.9\columnwidth]{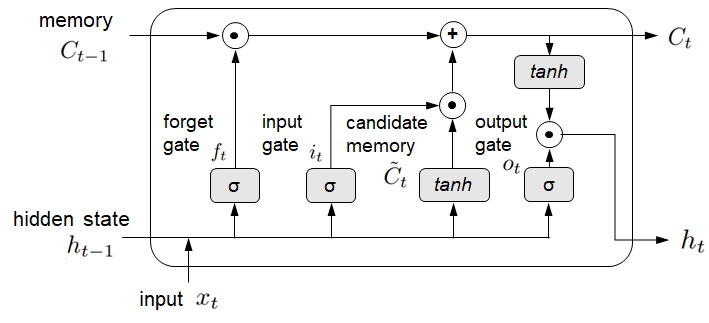}
\caption{Schematic diagram of LSTM module}\label{fig:lstm_schematic}
\end{figure}

The objective is to update the cell, $C_t$, over time using the input $x_t$ and the previous time step's hidden state, $h_{t-1}$. This process involves several key operations. First, a forget gate, $f_t$, selectively filters information from the past. Then, an input gate, $i_t$, regulates the amount of information from the candidate memory cell, $\Tilde{C}_t$, that should be incorporated into the current cell state, $C_t$. Finally, an output gate, $o_t$, governs the update of the hidden state, $h_t$. See Figure~\ref{fig:lstm_schematic}. The computations are represented as follows:

\begin{equation}\label{eqn:cellstate}
\begin{split}
    \Tilde{C}_t &= \tanh(W_{c}[{h}_{t-1},{x}_t] + b_{c})\\
    C_t &= f_t \odot C_{t-1} +i_t\odot \Tilde{C}_t\\
h_{t} &= o_{t}\odot \tanh(C_{t})\\
\end{split}
\end{equation}
The outputs from the forget gate, $f_{t}$, input gate, $i_{t}$, and output gate, $o_{t}$ are computed as shown below:
\begin{equation} \label{eqn:lstm_gates}
\begin{split}
f_t & = \sigma({W}_f[{h}_{t-1},{x}_t] + b_{f})\\
i_t & = \sigma({W}_i[{h}_{t-1},{x}_t] + b_{i})\\
o_t & = \sigma({W}_o[{h}_{t-1},{x}_t] + b_{o})\\
\end{split}
\end{equation}
Here, $\sigma$ and $\tanh$ represent non-linear activation functions, while $W_f$, $W_i$, $W_o$, and $W_{c}$ denote weight matrices corresponding to the forget gate, input gate, output gate, and candidate memory cell, respectively. Similarly, $b_f$, $b_i$, $b_o$, and $b_c$ represent the corresponding bias vectors.

\subsection{Uncertainty quantification}
Bayesian DL captures epistemic or aleatoric uncertainty by fitting probability distributions over the model parameters or model outputs, respectively. 
Epistemic uncertainty is modeled by placing a prior distribution over a model’s weights, and then calculating the posterior distribution of the weights given the training data, i.e. $P(w|\mathcal{D})$. During inference, the predictive probability of $\hat{y}$ for a test data $\hat{x}$ is computed by $P(\hat{y}|\hat{x}) = \mathbb{E}_{P(w|\mathcal{D})} P(\hat{y}|\hat{x},w)$. However, solving this is intractable. Popular methods use \textit{variational inference} \cite{blei2017} to obtain the posterior distribution, however these models are computationally expensive.  One such method is `Backpropagation by Bayes' \cite{blundell2015} which approximates the posterior distribution $P(w|\mathcal{D})$ with a simple distribution $q(w|\theta)$ parameterized by $\theta$ by minimizing their Kullback- Leibler (KL) divergence:
\begin{equation}
    \theta^* =  \arg\max_{\theta} \mathbf{KL}[q(w|\theta)||P(w|\mathcal{D})]
\end{equation}

In this study, we use \textit{dropout} -- a widely used regularization technique to perform the Bayesian inference by estimating the posterior distribution of weights of a NN \cite{gal2016, kendall2017}. Dropout is effective in preventing overfitting by randomly deactivating a fraction of neurons during each training iteration. This encourages the NN to become more robust and less dependent on specific neurons, thus mitigating overfitting to the training data.

To achieve Bayesian inference, we incorporate dropout at each layer of the NN during training. Additionally, we apply dropout during inference on the test data, enabling us to sample from the approximate posterior distribution. This approach, often referred to as Monte Carlo dropout, allows us to estimate uncertainty and obtain a more comprehensive understanding of model performance.
In this approach, $q(\mathbf{w}|\theta)$ is approximated as a mixture of two Gaussians with small variances and the mean of one of the Gaussians is fixed at zero. In a regression setting, the epistemic uncertainty is captured by the predictive variance of the model outputs. 
To the contrary, the aleatoric uncertainty is estimated by fitting a Gaussian distribution over the model output to capture the noise in data.
The model is trained to predict the mean, $\hat{y}$ and variance, $\sigma^2$. This results in the objective function being a likelihood minimization as given below.
\begin{equation}
    \mathcal{L(\theta)} = \dfrac{1}{D} \sum_{i}\dfrac{1}{2}\hat{\sigma}_{i}^{-2}||y-\hat{y}||^2 + \dfrac{1}{2}\log \hat{\sigma_{i}^2}
\end{equation}
where $D$ is the cardinality of data, $y$ and $\hat{y}$ are the true and predicted values respectively, and $\sigma^2$ is the prediction variance. For numerical stability, the model is trained to predict the natural logarithm of variance i.e., $s_i:=\log \hat{\sigma}_{i}^2$. Consequently, the objective function becomes,
\begin{equation}
    \mathcal{L(\theta)} = \dfrac{1}{D} \sum_{i}\dfrac{1}{2}\exp(-s_i)||y-\hat{y}||^2 + \dfrac{1}{2} s_i\label{eqn:objective}
\end{equation}

The total predictive uncertainty for $T$ stochastic forward passes of the model can be approximated as shown in Equation~\ref{eqn:totalunc}, where at each pass, a fixed fraction of neurons in each layer are randomly dropped. 
\begin{equation}
    \text{Var}(y) \approx {\underbrace{\textstyle \dfrac{1}{T} \sum_{t}^T \hat{y}^2 - \left( \dfrac{1}{T} \sum_{t}^T \hat{y} \right)^2}_{\mathclap{\text{Epistemic}}}}
 + {\underbrace{\textstyle \dfrac{1}{T} \sum_{t}^T \hat{\sigma}^2}_{\mathclap{\text{Aleatoric}}}} \label{eqn:totalunc}
\end{equation}


\section{Methodology}
In this section, we will define the model architectures used in our study, as well as the normalization schemes and transfer learning approaches that were used.

\subsection{Model training}
A stacked LSTM model with Bayesian inference by dropout is used for traffic flow prediction and uncertainty quantification in this study. Our model consists of three LSTM layers with 20, 20 and 10 units, followed by four dense layers with 10, 10, 6 and 2 units respectively, with LeakyReLU activation function \cite{maas2013rectifier} for dense layers respectively. 
In each layer, a fraction of hidden units are stochastically dropped during model training and inference by adding dropout. For each model, dropout rates of 2, 5 and 10 percent are used to evaluate its sensitivity on model accuracy. 
Specifically for the LSTM, we use \textit{recurrent dropout} \cite{semeniuta2016} which applies dropout to the cell update vector $\Tilde{C}_t$ (see Equation~\ref{eqn:LSTMdropout}) instead of dropping gate inputs \cite{gal2016rnn} or cell state \cite{moon2015}, which adversely affects its long-term memory. 
\begin{equation}\label{eqn:LSTMdropout}
    C_t = f_t \odot C_{t-1} +i_t\odot d(\Tilde{C}_t)
\end{equation}
where $d(\boldsymbol{\cdot})$ refers to the dropout mask. 

The models are trained and validated on 60\% and 15\% of the dataset, while the remaining 25\% is reserved for evaluating model performance. During the training process, we minimize the loss function (shown in Equation~\ref{eqn:objective}) for 100 epochs and choose the model with the lowest validation error. We use Adadelta \cite{adadelta} as the optimizer with the learning rate of 0.10, $\rho$ value of 0.95, and epsilon of 1e-7 to train the models. Once the models are trained, the stochastic forward pass is repeated for a finite number of runs (50 in our case) to obtain uncertainty estimates.


\subsection{Normalization techniques}
Normalization methods usually improve the process of NN training, particularly ensuring smoother gradients, faster training, and better generalization accuracy. 
The gradients of the weights in any layer are highly correlated to the outputs from the previous layer. Thus, changes in the output of one layer causes significant changes in summed inputs to the following layer. Batch normalization \cite{ioffe2015batch}, a technique that controls the mean and variance of inputs across mini-batches, has been shown to improve training efficiency for NNs. However, applying batch normalization to LSTM is more challenging, due the presence of a temporal dimension, where the hidden state of the network is dependent on the previous time steps. Applying batch normalization to the hidden state could result in the loss of temporal dependencies, which can negatively impact the model's performance. 

Alternatively, we investigate the use of spectral normalization as an alternative approach to enhance the generalizability of models. We also compare its performance with another normalization technique called layer normalization. A brief overview of both normalization techniques are presented below: 

\subsubsection{Layer normalization}
Layer normalization \cite{ba2016,xu2019} controls the mean and variance of the summed inputs to each layer of NN. For an input $x$ to the NN, the outputs $x^{l} = (x_1^l, x_2^l, \cdots, x_H^l)$ from layer $l$ is computed by:
\begin{equation}
  x^{l} = f^l(w^lx^{l-1} + b^l)
\end{equation}
for $l = 1, 2, \cdots, L$ where $f(\boldsymbol{\cdot})$ is an activation function, $w$ and $b$ represent the layer-specific weight matrix and bias vector, respectively. Here, $H$ represents the number of hidden units in layer $l$. Layer normalization re-centers and re-scales the vector $x^{l}$ using its mean and standard deviation, as shown below, before feeding to subsequent layers.

\begin{gather} 
\mu^l = \dfrac{1}{H}\sum_{i=1}^H x_i^{l} 
\text{  and   }
\sigma^l = \sqrt{\dfrac{1}{H}\sum_{i=1}^H (x_i^{l} - \mu^l)^2 + \epsilon}\\
LN_{\gamma\beta} \equiv \gamma \left(\dfrac{ x^{l} - \mu^l }{\sigma^l}\right) + \beta
\end{gather}
where $\gamma$ and $\beta$ are tunable scale and shift parameters, and $\epsilon$ is added for numerical stability. 

\subsubsection{Spectral normalization}
To stabilize NNs to data perturbation, spectral normalization \cite{miyato2018spectral} normalizes the weights of the NN.
For a NN, let 
$f_\Theta:x \rightarrow W_{\Theta}x+ b_{\Theta}$ be an affine transformation and $\xi$ be a perturbation vector with small $l_2$ norm, the response of the network to perturbation can be expressed as:
\begin{equation}
  \dfrac{\| f_{\Theta}(x+\xi) - f_{\Theta}(x)\|_2}{\| \xi\|_2} = \dfrac{\|W_{\Theta}\xi\|_2}{\|\xi\|_2} \leq \sigma(W_\Theta)
\end{equation}
where $\sigma(W_\Theta)$ is the spectral norm of the weight matrix $W_\Theta$, defined as its maximum singular value,
\begin{equation}
    \sigma(W_\Theta)= \max_{\xi\neq 0} \dfrac{\|W_{\Theta}\xi\|_2}{\|\xi\|_2}
\end{equation}
Further, decomposing the weight matrix $W_\Theta$ into layer specific weights and activation functions, it can be shown that, 
\begin{equation}
    \sigma(W_\Theta) \leq \prod_{l=1}^L \sigma(w^l)
\end{equation}

Therefore, by constraining the spectral norm of each weight matrix, $w^l$, model sensitivity to perturbation can be reduced. 
In contrast to spectral norm regularization \cite{yoshida2017}, which penalizes the spectral norm by adding an explicit regularization term to the loss function, the layer weights are simply divided by their corresponding spectral norm in our adopted approach.

\subsection{Transfer learning}
Transfer learning involves 'transferring' knowledge from a source domain to perform a similar task in another domain \cite{pan2009}. Transfer learning strategy eliminates the requirement to evaluate a new problem from scratch, rather the network is trained using data from the source domain and suitably adapted to new settings. This significantly reduces the learning time and amount of transfer-domain data required for training. Usually the discriminating layer in the NN is retrained using a small amount of data from the transfer domain, if available. In our case, we retrain the dense layers of the NN, while preserving the LSTM layers, to adapt the model to perform traffic prediction at the transfer location.

\section{Data}
The data used in this study for model training and evaluation is obtained from the California Department of Transportation's Performance Measurement System (PeMS), which is popularly used for traffic data modeling. Traffic data (flow, occupancy, speed) are recorded using vehicle detector sensors (VDS) along the freeways and ramps. Raw data sampled at 30 second intervals are aggregated every 5 minutes. 
Flow data for one year (i.e., 104942 samples) from VDS 1114805 on California Interstate-05 NB in District 11 were used for training and testing of the models. The performance of the models are evaluated on 25\% of the dataset that was never used in training. However, since both training and test data share the same distribution characteristics, evaluating model performance only on the test split of the dataset only ensures model robustness within the bounds of the data distribution. The generalizability of the model is therefore additionally evaluated on a set of out-of-distribution datasets. 

\begin{figure}[htb!]
\centering
\includegraphics[width=1\columnwidth]{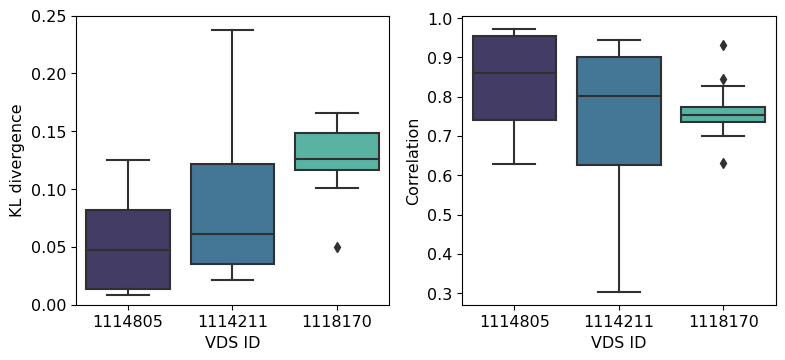}
\caption{Comparison of distribution of KL divergence and correlation between training and out-of-distribution datasets}
\label{fig:kl_cor}
\end{figure}

\begin{figure}[htb!]
\centering
\includegraphics[width=1\columnwidth]{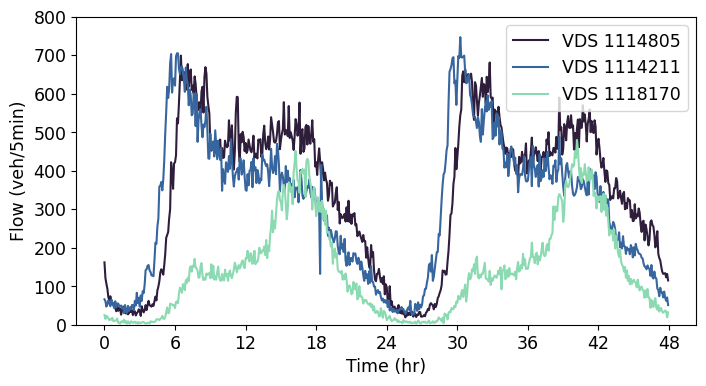}
\caption{Traffic flow patterns in training and out-of-distribution datasets for 48 hours}
\label{fig:flow_data}
\end{figure}

To identify the out-of-distribution datasets, different detector locations within District 11 of California were compared. The similarity of traffic flow at these other locations were evaluated using two techniques: KL divergence and correlation analysis. KL divergence was used to compare the probability distributions of traffic flow between different stations, while correlation analysis was used to measure the strength and direction of the linear relationship between traffic data from multiple stations. The KL divergence of daily traffic flows from multiple detector locations was computed with respect to the training station (i.e., VDS 1114805) over a period of 30 days. Multiple days of data were used to account for the stochasticity in daily traffic patterns. We then ranked the stations based on the median KL divergence and selected two stations: (1) one with the highest KL divergence, indicating the most distinct traffic pattern and (2) another with a KL divergence in between the training station and that of the station with highest KL divergence. A similar analysis was conducted using the correlation as a metric and resulted in the selection of the same two locations. The similarity scores for the training and out-of-distribution datasets are presented in Figure~\ref{fig:kl_cor}. The traffic flow patterns at these three locations over a period of 2 days is shown in Figure~\ref{fig:flow_data} to illustrate the existence of different patterns of traffic. 

\section{Results} 
In this study, three DL models -- 1) regular Bayesian LSTM without normalization, 2) Bayesian LSTM with layer normalization, and 3) Bayesian LSTM with spectral normalization are considered for uncertainty quantification in traffic flow prediction. 
Stochastic dropout is used for all models to estimate the prediction mean and total uncertainty (given by total standard deviation, see Equation~\ref{eqn:totalunc}) through finite number of model runs. To gain a deeper understanding of the impact of normalization, the total uncertainty is further decomposed into aleatoric and epistemic uncertainties.

The performance evaluation of the models includes assessing the accuracy of predictions and the level of uncertainty. We evaluate the model prediction performances by comparing the prediction mean with the corresponding true flow values using three metrics: root mean squared error (RMSE), mean absolute percentage error (MAPE) and $R^2$ as defined below.

\begin{equation} 
\mathrm{RMSE}={\sqrt{\dfrac{1}{n}\sum_{i=1}^{N} \left [y_{i}-\hat {y_{i}}\right]^{2}} }\label{eqn:rmse}
\end{equation}

\begin{equation} 
\mathrm{MAPE}=\dfrac{1}{N}\sum_{i=1}^{N} \lvert{ \dfrac{y_{i}-\hat {y_{i}}}{y_{i}} }\rvert \label{eqn:mape}
\end{equation}

\begin{equation} 
\mathrm{R^2}=1-\dfrac{\sum_{i=1}^{N} \left [y_{i}-\hat {y_{i}}\right]^{2}}{\sum_{i=1}^{N} \left [y_{i}-\bar {y_{i}}\right]^{2}} \label{eqn:mae}
\end{equation}

where $y_i$ represents the 'ground truth' or true value of the observation $i$, $\hat{y_i}$ is the predicted value of $y_i$ for $i=1, 2,\dots T$. 

\subsection{Performance on training dataset}

\begin{figure*}[h]
\centering
\includegraphics[width=1\linewidth]{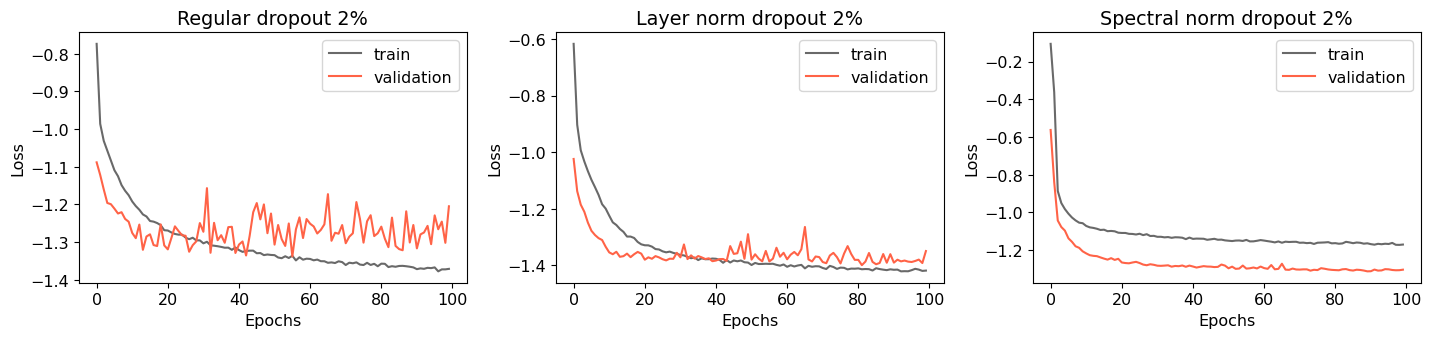}
\caption{Training loss evolution (a) Bayesian LSTM without normalization (b) Bayesian LSTM with layer-normalization (c) Bayesian LSTM with spectral normalization with dropout rate of 2\%}
\label{fig:train_val_loss}
\end{figure*}

First, we present the evolution of training and validation losses for the models during the training process corresponding to a dropout rate of 2\% in Figure~\ref{fig:train_val_loss}. 
For the regular dropout model without any normalization, we observed that both the training and validation losses initially decreased as the model was trained, indicating that the model was effectively learning from the training data and generalizing well to unseen data i.e., the validation split. However, after a certain number of epochs the validation loss starts to increase. Therefore, the model was not generalizing well to the validation data and was instead memorizing the training data - a phenomenon referred to as `overfitting'. This indicates that the model is becoming too complex and may be fitting the model to noise in the training data. 

It is also important to analyze the rate at which the validation loss increases because a faster rate of increase indicates that the model is overfitting more quickly, which means that the model's generalizability to unseen data may be severely limited. Bayesian LSTM with layer-normalization still shows some signs of overfitting, however, this happens at a much lower rate compared to the case when no normalization were used. By controlling the mean and standard deviations of inputs to different layers of NN, the layer-normalization allowed the model to focus more on important features and patterns in the data, rather than fitting noise in the training data. Hence, the model's performance on unseen data can be improved with layer-normalization by reducing overfitting, while still maintaining a high level of performance on the training data. 

In contrast, when applying the spectral normalization method, both the training and validation losses exhibited decreasing trends. This indicates that the model was not overfitting and suggests that the normalization technique effectively reduced the complexity of the model
By controlling the spectral norms of the weight matrices in the NN, this method encouraged the model to learn simpler and more generalized representations of the data. This, in turn, can lead to improved generalization performance on new data. 
However, due to over-regularization, both the training and validation losses for the spectral normalized model were higher compared to the regular model and the layer-normalized model. 
Specifically, the validation losses for the regular, layer-normalized, and spectral normalized models were -1.3382, -1.3990, and 1.3122 for 2\% dropout. Additionally, the experiments were performed for dropout rates of 5\% and 10\%, however the results are only presented corresponding to 2\% dropout for brevity. The results and conclusions look similar for the other dropout rates, however, the prediction accuracy decreases and overfitting increases for larger dropout rates.

\begin{figure*}[]
\centering
\includegraphics[width=1\linewidth]{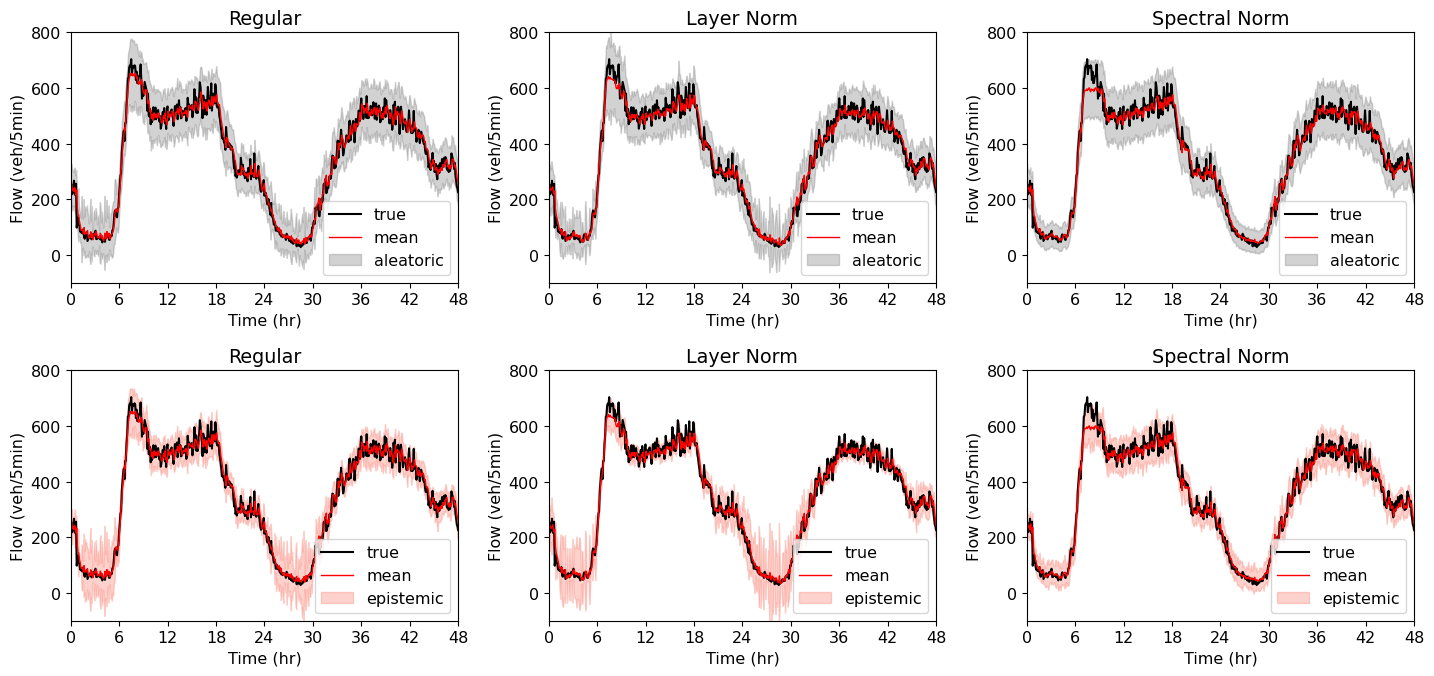}
\caption{(a) Aleatoric and (b) epistemic uncertainty on in-distribution test data for regular, layer normalized and spectral normalized models with 2\% dropout rate}
\label{fig:unc_test_d2}
\end{figure*}

In the context of evaluating the prediction performance of each model on the test data, the mean prediction was calculated. The results are presented in Table~\ref{tab:perf_test}, which includes the RMSE, $R^2$, and MAPE metrics. Specifically, the regular model exhibited the lowest RMSE, however the layer- and spectral normalization did not increase the RMSE much. The general predictions are fairly accurate -- the mean prediction (red curve) accurately follows the real traffic flows (black curve), see Figure~\ref{fig:unc_test_d2}. Further, the RMSE and $R^2$ values follow similar trends as the validation loss obtained for each model. This behavior can be attributed to the fact that the validation loss, as defined in Equation~\ref{eqn:objective}, also takes into account the squared deviations between the predicted and true values.

\begin{table}[]
\caption{Prediction performance comparison of models on test data}\label{tab:perf_test}	
\begin{center}
\begin{tabular}{lccc}
\hline
Metric & Regular & LN & SN \\ \hline
RMSE   & 29.7120  & 29.8062    & 30.9620       \\ 
$R^2$  & 0.9775   & 0.9774     & 0.9756        \\ 
MAPE   & 0.1111   & 0.1071     & 0.1144        \\ \hline
\end{tabular}
\end{center}
\end{table}

\begin{figure}[t]
\centering
\includegraphics[width=1\columnwidth]{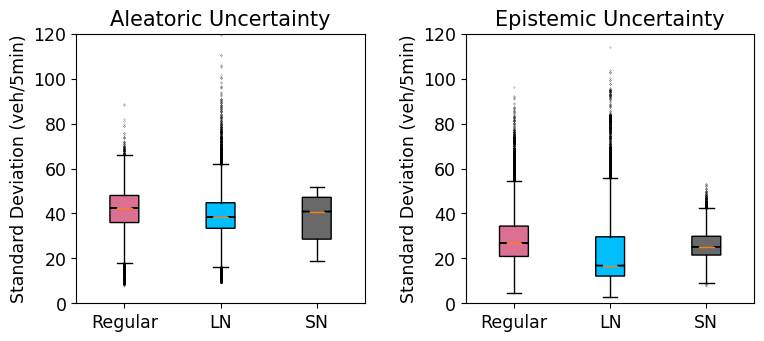}
\caption{Comparing aleatoric and epistemic uncertainty across models in-distribution test data}
\label{fig:unc_test_compare}
\end{figure}

Next, the uncertainty estimates are compared. Figure~\ref{fig:unc_test_d2} shows the 95th percentile confidence intervals of the estimates of the aleatoric and epistemic uncertainties using grey and red bands in each subplot. As depicted, there are significant differences in the uncertainty estimates among the three models. 
To better understand how the overall aleatoric and epistemic uncertainty of each model compares, Figure~\ref{fig:unc_test_compare} plots the distribution of the magnitude of the uncertainties estimated over the test data. In these box-plots, the colored boxes represent the boundaries for the 25th and 75th percentile of the standard deviations, with the notch inside the box representing the median. The whiskers extend from the box by 1.5 times the inter-quartile range and flier points are those past the end of the whiskers.
The figures illustrate that, although the aleatoric uncertainty estimates in layer normalized model exhibit a high degree of dispersion, the median standard deviation is lower than the regular model. Similarly, the layer normalized model outperforms the regular model in terms of epistemic uncertainty. Nevertheless, in both cases, the spectral normalization approach effectively controls the standard deviation.

\begin{figure*}[]
\centering
\includegraphics[width=0.80\linewidth]{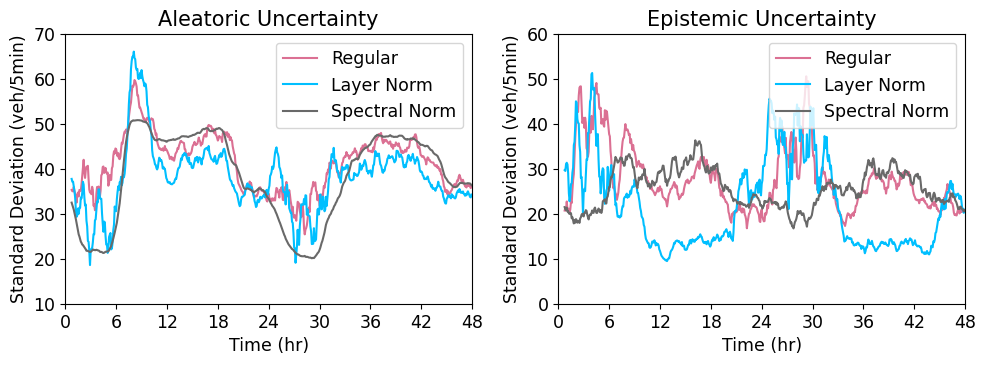}
\caption{Comparison of temporal variation of (a) aleatoric and (b) epistemic uncertainties for different models}
\label{fig:unc_variation}
\end{figure*}

The variation in the magnitude of the estimated aleatoric and epistemic uncertainties over time are shown in Figure~\ref{fig:unc_variation}.
The aleatoric uncertainty estimates follow the general trend of the data for all models, with lower uncertainty observed during periods characterized by lower flow rates and and greater uncertainty observed during periods of higher flows, such as the peak congestion. This observation can be attributed to the chaotic and unpredictable behavior of the system during congested states, which are commonly associated with higher flows. 

However, minor differences are observed for regular and layer normalized models, where they show a slight increase in aleatoric uncertainty during low flow regions (0 to 6 hr and 24 to 30 hr) and morning peak congestion period (~ 8 hr), compared to other flow regimes. This may be attributed to the limited knowledge of the models about the low flow and peak congestion regimes, as the training data only consisted of approximately 20\% and 6\% of data belonging to these regimes. 

This effect is particularly evident in the epistemic uncertainty, where we observe a more pronounced higher standard deviation specifically for low flow regimes in these two models. In contrast, the spectral normalized model was able to capture trends despite the limited representation of low flow data in the training dataset. This suggests that the spectral normalization approach may be more effective at capturing and generalizing to unseen datasets, compared to the other models. 
However, for regimes with higher data representation, the layer normalized model performs better than spectral norm model - therefore provides lower epistemic uncertainty estimates.

\subsection{Performance on out-of-distribution datasets}

Here, we aim to investigate the impact of normalization on the generalizability of the models by evaluating their performance, both in terms of accuracy and uncertainty, on out-of-distribution datasets, namely VDS 1114211 and VDS 1118170. Initially, we assess the performance of the trained models on these datasets without any additional training, providing insights into how well the models can generalize to data that differs from the training distribution. Subsequently, we proceed to retrain the models using partial data from these out-of-distribution datasets. Finally, we re-evaluate the models' performance to determine the impact of retraining on their ability to handle these datasets.

The prediction accuracy of the models in transfer learning without and with retraining on out-of-distribution datasets is shown in Table~\ref{tab:perf_od_d2} corresponding to dropout rate of 2\%.
We observe that the performance of models at a new station is not the same as that at the training station. This is expected since the models are being tested on samples that were not encountered during training. We observe the regular model to exhibit the highest $R^2$ and least RMSE value compared to both layer and spectral-norm models. Similar trends hold true for other dropout values, and hence omitted for brevity. However, it was observed that as the dropout rate increases, models exhibits higher regularization, resulting in a decrease in prediction accuracy. 

\begin{figure}[t]
\centering
\includegraphics[width=0.95\columnwidth]{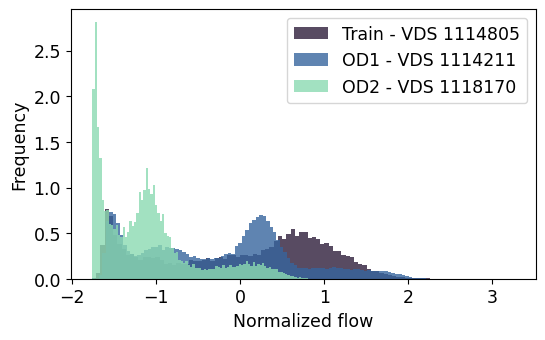}
\caption{Comparing the distribution of the normalized flow values in the out-of-distribution datasets with that in the training data}
\label{fig:data_distribution}
\end{figure}

Surprisingly, even when faced with significantly different temporal patterns in these out-of-distribution datasets, we observe acceptable levels of 
prediction performance -- RMSE values are not observed to increase. 
Interestingly, the model's performance is contingent upon the underlying data distribution, rather than the temporal pattern itself. For instance, encountering previously unseen values can lead to errors or uncertainties in the model's predictions. To further comprehend this phenomenon, we can refer to Figure~\ref{fig:data_distribution}, which illustrates a comparison between the distribution of the normalized flow values from out-of-distribution datasets and the training data. 

\begin{figure*}[t]
\centering
\includegraphics[width=1\textwidth]{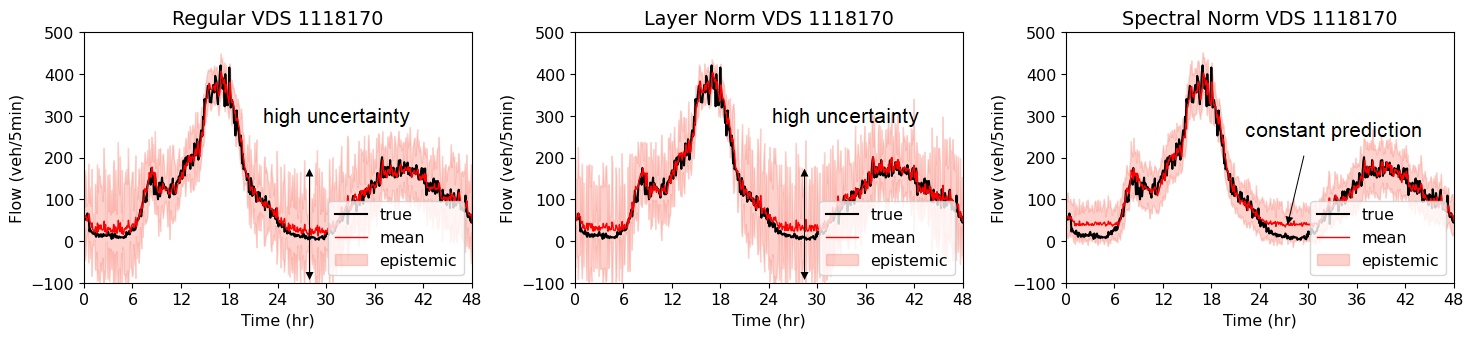}
\caption{Comparing the performance of models on the out-of-distribution dataset VDS 1118170}.
\label{fig:spectral_constant}
\end{figure*}

A sample of model predictions over a 48 hour period on one of the out-of-distribution datasets, namely VDS 1118170 without retraining is shown in Figure~\ref{fig:spectral_constant}. 
It can be seen that the models' predictions in low flow regimes tend to be characterized by errors and uncertainty. This is likely due to the data distribution of VDS 1118170 exhibiting a high skewness with an abundance of low flow values that were not encountered during the model's training phase. 
Interestingly, the spectral normalized model provides a constant prediction for the low flow regimes, unlike the regular or layer normalized models, see Figure~\ref{fig:spectral_constant}. The spectral normalized model treats extreme low flows as perturbations to the low flow values observed during training and therefore restricts its predictions to a constant value. However, it is worth noting that all models performs remarkably well in other flow regimes, resulting in low RMSE values within the overlapping regions of the distributions. 
Note that from a traffic prediction perspective, errors in the low flow regime are less critical since congestion management strategies are targeted for high flow regimes.  

In contrast, the distribution of VDS 1114211 closely resembles the training distribution, suggesting that the model has encountered similar flow conditions during training, resulting in accurate predictions and low RMSE values.

When the dense layers (and layer normalization parameters for the layer normalized model) of the models are retrained using 20\% of the data from each of these locations, the RMSE values tend to decrease in majority of cases. However, it is important to note that the MAPE often increases in few cases. The reason behind this behavior lies in the objective of transfer learning, which aims to minimize the loss function described in Equation~\ref{eqn:objective}, primarily based on square error loss, and the sensitivity of MAPE to errors in low flow regions, which might be affected during the retraining process.

\begin{table*}[]
\caption{Prediction performance comparison of models on out-of-distribution test data with 2\% dropout rate}\label{tab:perf_od_d2}	
\begin{center}
\begin{tabular}{cccccc}
\hline
{Dataset} & {Metric} & {TL} & {Regular} & {LN} & {SN} \\ \hline
\multirow{6}{*}{VDS 1114211} & \multirow{2}{*}{{RMSE}} & No & 25.9613 & 26.6853 & 28.3968 \\ \cline{3-6} 
 &  & Yes & 25.9529 & 25.5755 & 27.7584 \\ \cline{2-6} 
 & \multirow{2}{*}{{$R^2$}} & No & 0.9765 & 0.9752 & 0.9719 \\ \cline{3-6} 
 &  & Yes & 0.9765 & 0.9772 & 0.9731 \\ \cline{2-6} 
 & \multirow{2}{*}{{MAPE}} & No & 0.0967 & 0.0962 & 0.0964 \\ \cline{3-6} 
 &  & Yes & 0.1065 & 0.0990 & 0.1017 \\ \hline
\multirow{6}{*}{VDS 1118170} & \multirow{2}{*}{{RMSE}} & No & 19.5398 & 21.4862 & 22.9500 \\ \cline{3-6} 
 &  & Yes & 19.5259 & 18.8665 & 20.7542 \\ \cline{2-6} 
 & \multirow{2}{*}{{$R^2$}} & No & 0.9683 & 0.9617 & 0.9563 \\ \cline{3-6} 
 &  & Yes & 0.9683 & 0.9704 & 0.9642 \\ \cline{2-6} 
 & \multirow{2}{*}{{MAPE}} & No & 0.5646 & 0.7660 & 0.9967 \\ \cline{3-6} 
 &  & Yes & 0.4975 & 0.4170 & 0.6957 \\ \hline

\end{tabular}
\end{center}
\end{table*}

\begin{figure*}[]
\centering
\includegraphics[width=0.6\textwidth]{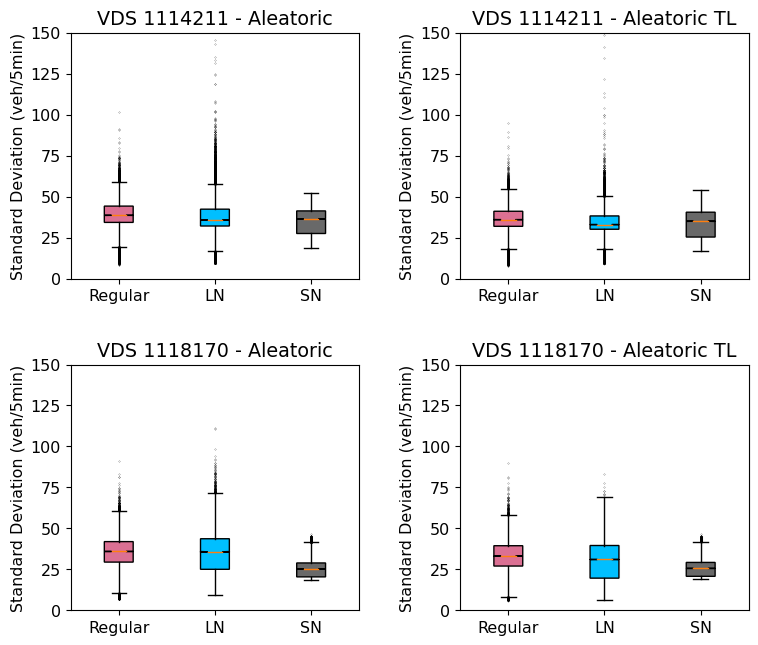}
\caption{Aleatoric uncertainty on out-of-distribution datasets for models with 2\% dropout}
\label{fig:unc_OD_alea_d2}
\end{figure*}

\begin{figure*}[]
\centering
\includegraphics[width=0.6\textwidth]{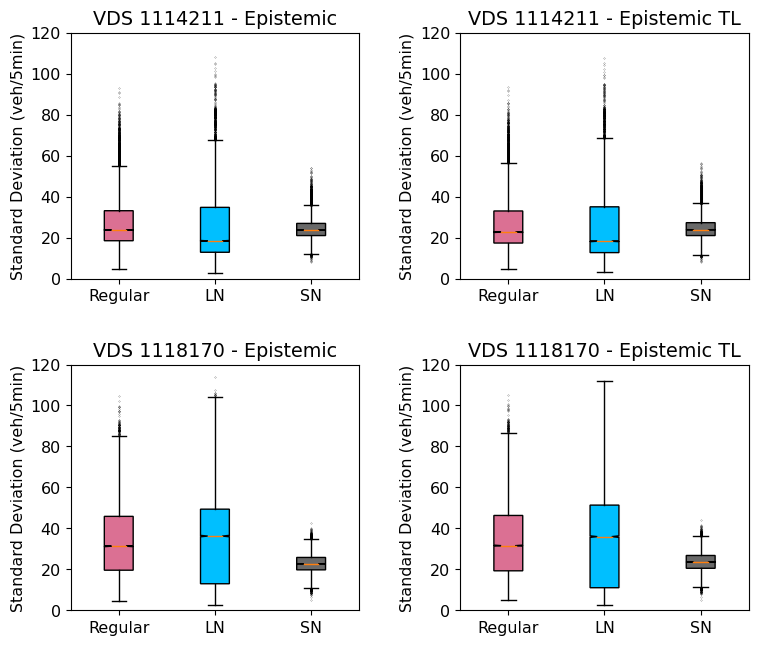}
\caption{Epistemic uncertainty on out-of-distribution datasets for models with 2\% dropout}
\label{fig:unc_OD_epis_d2}
\end{figure*}

Normalization is also observed to have a significant impact on uncertainty estimation. The uncertainty plots presented in Figures~\ref{fig:unc_OD_alea_d2} and \ref{fig:unc_OD_epis_d2} demonstrate the aleatoric and epistemic uncertainty at the two test locations, both before and after retraining. Consistently with the trends observed in the training data, we find that normalization reduces both aleatoric and epistemic uncertainty in these plots. Notably, spectral normalization exhibits a greater reduction in uncertainty compared to layer normalization. This advantage becomes more pronounced as the dissimilarity between the out-of-distribution datasets increases. Despite the marginally higher RMSE obtained with spectral normalization, one might still prefer to utilize it due to its ability to provide more confident predictions. The reduced uncertainty associated with spectral normalization enhances the model's reliability and provides a higher level of confidence in its predictions, which can be crucial in certain applications.

Furthermore, it is important to consider that aleatoric uncertainty estimates obtained on these datasets without retraining might not accurately reflect their true representation. This is because the estimated data variance ($\sigma^2$) is learned solely from the training data and remains unchanged. However, after retraining, the models exhibit a more precise estimation of data uncertainty as they adapt to the characteristics of the new datasets. 
Notably, for the spectral normalized model, the standard deviation increased for both stations, accounting for changes in flow patterns between the training and out-of-distribution datasets. On the other hand, the other models did not demonstrate significant changes in the aleatoric uncertainty estimates. Regarding epistemic uncertainty, there were no notable changes observed across the models. This observation suggests that the regular and layer normalized models tend to be overconfident in their predictions, possibly due to their inability to generalize well to out-of-distribution datasets. In contrast, the spectral normalized model exhibits a more realistic estimation of uncertainty, indicating its ability to provide reliable and generalizable predictions.

\subsection{Model Interpretation}
To gain a deeper understanding of the model performances, traffic time series were labeled into four regimes: 1) low flow (0 to 6 hr), 2) increasing flow (6 to 8 hr), 3) high flow (8 - 18 hr) or congestion and 4) decreasing flow (18 to 24 hr) which are appropriately color-coded as shown below. By comparing the gradients and feature space outputs of each model across these flow regimes, we aim to discern their specific characteristics and differences.

\begin{figure}[!]
\centering
\includegraphics[width=1\columnwidth]{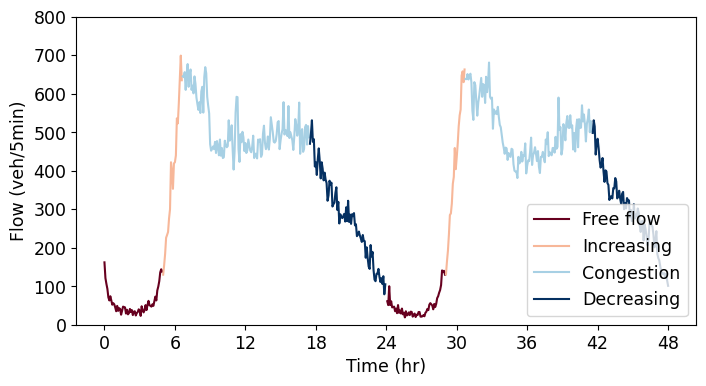}
\caption{Flow segmented based on regimes}
\end{figure}

\begin{figure*}[!]
\centering
\includegraphics[width=0.95\textwidth]{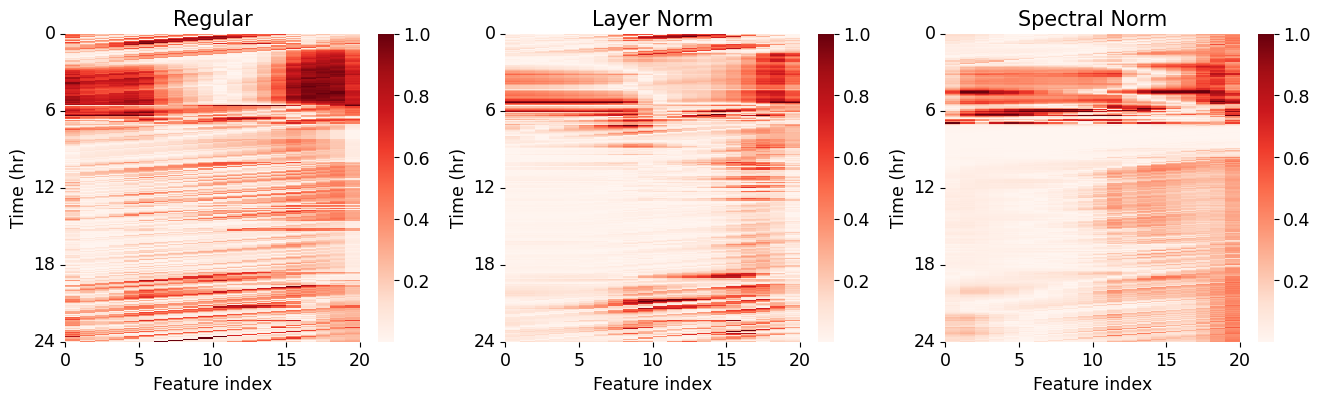}
\caption{Comparison of gradients of regular, layer normalized and spectral normalized models for 2\% dropout}
\label{fig:gradients}
\end{figure*}

Figure~\ref{fig:gradients} shows the color-maps of gradients normalized across time for the three models corresponding to different input vectors over a period of 24 hours (i.e. 288 samples in intervals of 5 minutes). Note that the x-axis represents the feature index, which is the historical flow data from the previous time steps with the latest flow data appearing at the right-hand-side end, whereas the y-axis corresponds to different times of the day. The model gradients are compared to identify patterns that could lead to higher generalizability of the spectral normalized models.

The regular models exhibit higher gradients, indicating increased sensitivity of the model outputs to historical traffic flows, particularly during the low flow regime (0 to 6 hr) and the increasing flow regime (6 to 8 hr). The layer normalized models demonstrate slightly lower gradients compared to the regular models, while the spectral normalized models exhibit the lowest gradients.
The increased sensitivity to specific data patterns in regular and layer normalized models leads to larger variations in predictions and, consequently, wider uncertainty bands. On the other hand, the presence of added regularization in spectral normalization allows the model to focus more on generic trends rather than specific data characteristics. As a result, the spectral normalized model demonstrates relatively consistent gradient characteristics throughout the day, showing less sensitivity to different inputs across traffic regimes. 

\begin{figure*}[!]
\centering
\includegraphics[width=0.95\textwidth]{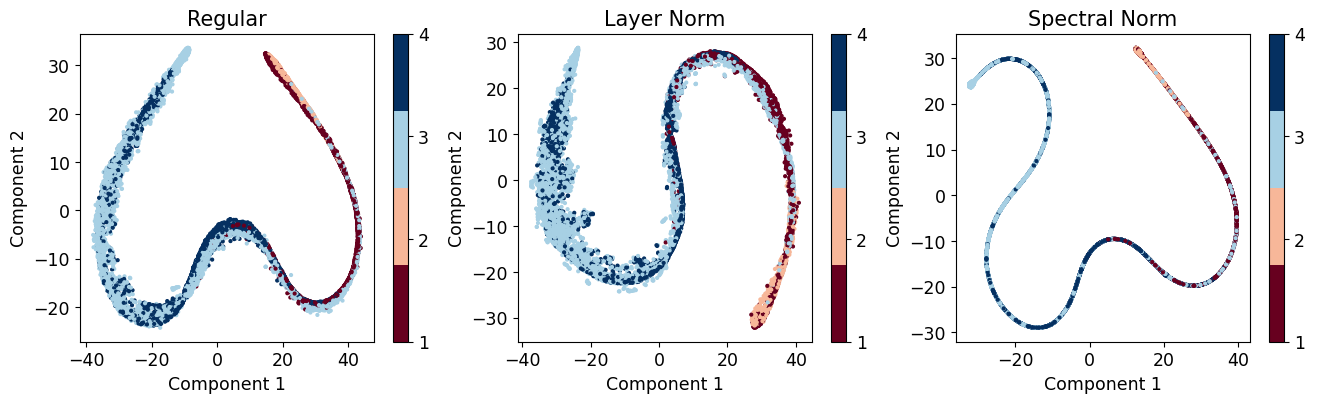}
\caption{Feature space visualizations of layer outputs of models}
\label{fig:feature}
\end{figure*}

Further, feature-space representations of the outputs from the penultimate layers of the models were visually compared in Figure~\ref{fig:feature} to understand their relation with the gradients explored before. Based on the model configuration, the six dimensional feature outputs were suitably reduced to two dimensions using a non-linear dimension reduction technique, $t$-Stochastic Neighbor Embedding \cite{van2008}. 

We have observed that the outputs from the regular dropout model exhibit high dispersion in feature space for data within the same traffic states, indicating sensitivity to specific data and consequently, limited generalizability. 
Particularly, we find that the feature outputs demonstrate significant dispersion in space for low- and increasing flow states, which aligns with the higher gradients observed in these regimes. 
In Table~\ref{tab:tsne}, we compare the variances of outputs belonging to specific traffic regimes in the 6-dimensional feature space for each model. 
Similar trends are observed for the layer and spectral normalized models, although these model outputs show localized behavior and less sensitivity. A notable improvement in feature space localization is observed for the spectral normalized model. This enhanced localization of traffic data from different flow regimes in the feature space has important implications for the model's learning process and contributes to the stability and improved generalizability of the spectral normalized model. This characteristic allows the model to handle data perturbations more effectively and provides increased robustness and reliability in capturing different traffic patterns.

These trends hold true across various out-of-distribution datasets, although we have omitted the detailed results for brevity. This aspect is particularly valuable in scenarios where it is not feasible to train models specifically for a given location due to limited data availability. In such cases, models trained and validated on abundant data can be effectively transferred for use in different domains, such as a station with distinct characteristics. 

\begin{table}[]
\caption{Variance of 6-dimensional outputs of models with 2\% dropout for different traffic flow regimes}\label{tab:tsne}
\begin{center}
\begin{tabular}{lccc}
\hline
Regime     & Regular & LN & SN \\ \hline
Low        & 1.1304   & 1.0363     & 0.4723\\ 
Increasing &  1.8117   & 1.5464     & 0.8096\\
High       & 0.6532   & 0.7374    & 0.2836\\
Decreasing & 0.4890   & 0.6063    & 0.1845\\ \hline
\end{tabular}
\end{center}
\end{table}

\section{Conclusions} 
Training deep learning (DL) models for traffic prediction that exhibit high generalizability across different datasets is of significant importance. While the objective of traffic management is to predict future traffic states at various locations, it is often not feasible to train location-specific prediction models due to limited data availability or budget constraints. Therefore, traffic prediction models with higher generalizability could be used to predict traffic conditions at locations where data is scarce.
Moreover, a significant drawback of many existing approaches is their failure to provide forecasts with uncertainty estimates, which are essential for effective traffic operations and control. In this study, a spectral normalized Bayesian LSTM is proposed for uncertainty estimation in traffic flow prediction, aiming to achieve higher generalizability. The performance of this model is compared with other baselines, including a model without normalization and one with layer normalization. 
The robustness and generalizability of the proposed model were evaluated using flow data with varying degrees of dissimilarity with the training data. The results suggest that the spectral normalized model offers considerably lower uncertainty estimates and generalizes better for unseen datasets compared to the other baseline models considered.
However, the prediction performance of the spectral normalized model is sometimes marginally compromised due to the added regularization inherent in spectral normalization. The improved generalizability of the spectral normalized model can be attributed to its enhanced resilience to data perturbation, due to an upper bound on performance drop caused by such perturbations. As a result, spectral normalized models trained and validated on data from one location can be suitably transferred for use in a different domain with varying traffic patterns.
Overall, the enhanced generalizability of the spectral normalized models offers promising opportunities for deploying models across diverse traffic scenarios, ensuring reliable predictions, and minimizing the need for extensive training data at every specific location.

\bibliography{manuscript.bib}

\end{document}